\documentclass[10pt,twocolumn,letterpaper]{article}

\usepackage{iccv}              
\usepackage[accsupp]{axessibility}
\usepackage{bm}
\usepackage{amssymb}
\usepackage{amsmath}
\usepackage{multirow}
\usepackage{booktabs}
\usepackage{array}
\usepackage{threeparttable}

\definecolor{iccvblue}{rgb}{0.21,0.49,0.74}
\usepackage[pagebackref,breaklinks,colorlinks,allcolors=iccvblue]{hyperref}

\def\paperID{9307} 
\def\confName{ICCV}
\def\confYear{2025}

\title{Prompt Guidance and Human Proximal Perception for HOT Prediction with Regional Joint Loss}

\author{Yuxiao Wang$^{1}$\quad
Yu Lei$^{2}$\quad
Zhenao Wei$^{1}$\quad
Weiying Xue$^{1}$
\\
Xinyu Jiang$^{1}$\quad
Nan Zhuang$^{3}$\quad
Qi Liu$^{1}$\thanks{Corresponding author}
\\
$^{1}$South China University of Technology \quad
$^{2}$Southwest Jiaotong University \quad
$^{3}$Zhejiang University
\\
{\tt\small \{ftwangyuxiao, drliuqi\}@scut.edu.cn}
}

\begin{document}
\maketitle
\begin{abstract}
The task of Human-Object conTact (HOT) detection involves identifying the specific areas of the human body that are touching objects.
Nevertheless, current models are restricted to just one type of image, often leading to too much segmentation in areas with little interaction, and struggling to maintain category consistency within specific regions.
To tackle this issue, a HOT framework, termed \textbf{P3HOT}, is proposed, which blends \textbf{P}rompt guidance and human \textbf{P}roximal \textbf{P}erception. 
To begin with, we utilize a semantic-driven prompt mechanism to direct the network's attention towards the relevant regions based on the correlation between image and text.
Then a human proximal perception mechanism is employed to dynamically perceive key depth range around the human, using learnable parameters to effectively eliminate regions where interactions are not expected.
Calculating depth resolves the uncertainty of the overlap between humans and objects in a 2D perspective, providing a quasi-3D viewpoint.
Moreover, a Regional Joint Loss (RJLoss) has been created as a new loss to inhibit abnormal categories in the same area. A new evaluation metric called ``AD-Acc.'' is introduced to address the shortcomings of existing methods in addressing negative samples.
Comprehensive experimental results demonstrate that our approach achieves state-of-the-art performance in four metrics across two benchmark datasets. Specifically, our model achieves an improvement of \textbf{0.7}$\uparrow$, \textbf{2.0}$\uparrow$, \textbf{1.6}$\uparrow$, and \textbf{11.0}$\uparrow$ in SC-Acc., mIoU, wIoU, and AD-Acc. metrics, respectively, on the HOT-Annotated dataset. The sources code are available at https://github.com/YuxiaoWang-AI/P3HOT.
\end{abstract}    
\begin{figure}[ht]
\centering 
\includegraphics[width=\linewidth]{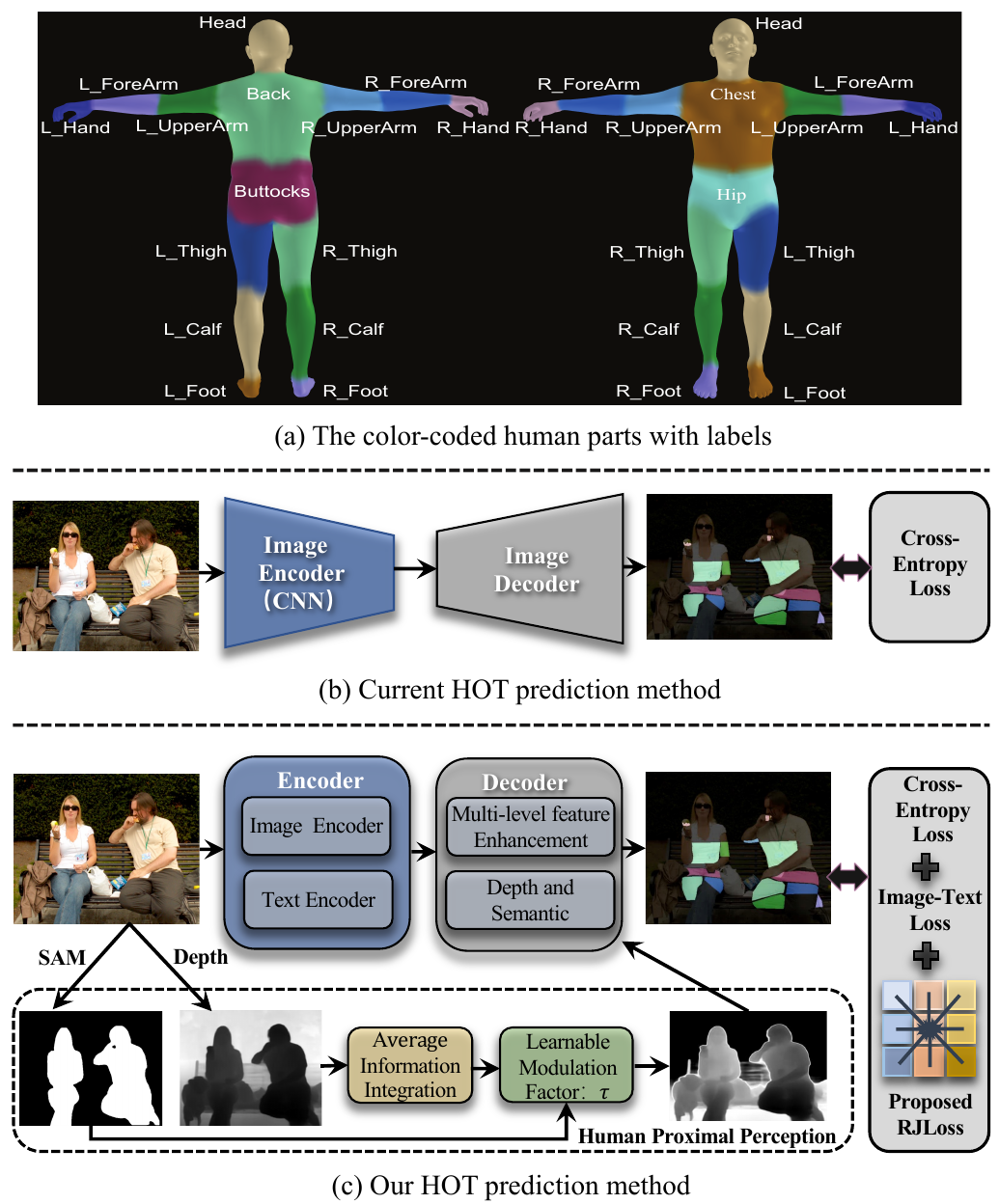}
\caption{Current method vs Ours.}
\label{fig:figure_1}
\vspace{-0.5cm}
\end{figure}

\section{Introduction}
\label{sec:intro}
Human-Object conTact (HOT) prediction~\cite{hot} originated from Human-Object Interaction (HOI) detection, an emerging advanced semantic understanding task ~\cite{wang2024ted,wang2024freea,wang2024review}. HOI focuses on detecting humans, objects, and their interactions. It does not specify which parts of the human body come into contact with objects. However, identifying specific contact points on the human is required in the HOT task. This technology can be applied in various fields such as human-computer interaction~\cite{wang2024ted}, virtual reality~\cite{illahi2023learning}, and gesture recognition~\cite{cui2023hand}.


HOT divides into 18 categories (including background)
, as shown in Figure \ref{fig:figure_1}(a). 
Chen et al.~\cite{hot} introduced DHOT, which employed ResNet-50 as feature extraction network and designed a contact attention mechanism coupled with human masks for HOT prediction (Figure \ref{fig:figure_1}(b)). 
Nonetheless, DHOT only utilized features from the last layer of the feature extraction network, which can be disadvantageous for segmentation tasks that usually rely on low-level features to improve segmentation accuracy~\cite{cui2023hand}. They predicted HOT from a single image modality, neglecting the guidance provided by other modalities. Additionally, the DHOT method failed to consider the consistency of classes within a region, leading to predictions that may include other classes within a particular class.


To overcome these challenges, we propose prompt guidance and human proximal perception method for HOT Prediction (a simplified diagram is shown in Figure \ref{fig:figure_1}(c)). Specifically, a text prompt mechanism is introduced to focus on human contact parts within the feature map. A human proximal perception (HPP) module is designed that utilizes a learnable parameter and human masks to dynamically perceive the depth range around the human. Furthermore, to regain initial texture detail loss during downsampling in the feature extraction stage, 
the lost image texture information is continuously refined by integrating outputs from each block of the feature extraction into the upsampling operations of the decoder. Finally, a novel loss function is designed, termed RJLoss, to ensure category consistency within regions and reduce the impact of abnormal categories.
Our contributions are summarized as follows:

\begin{itemize}
    \item We are the first to integrate textual information into HOT, setting the stage for future research into multi-modal HOT. 
    \item The proposed HPP module is leveraged from a pseudo-3D perspective to assist HOT. A learnable threshold is used to dynamically perceive areas where interaction may exist, effectively reducing unnecessary interference.
    \item A novel loss function, i.e., RJLoss, is introduced to ensure intra-region category consistency. A new evaluation metric is suggested to overcome the shortcomings of current metrics when accounting for incorrect predictions.
\end{itemize}

\section{Related work}
\label{sec:related_work}

\begin{figure*}[ht]
\centering 
\includegraphics[width=\linewidth]{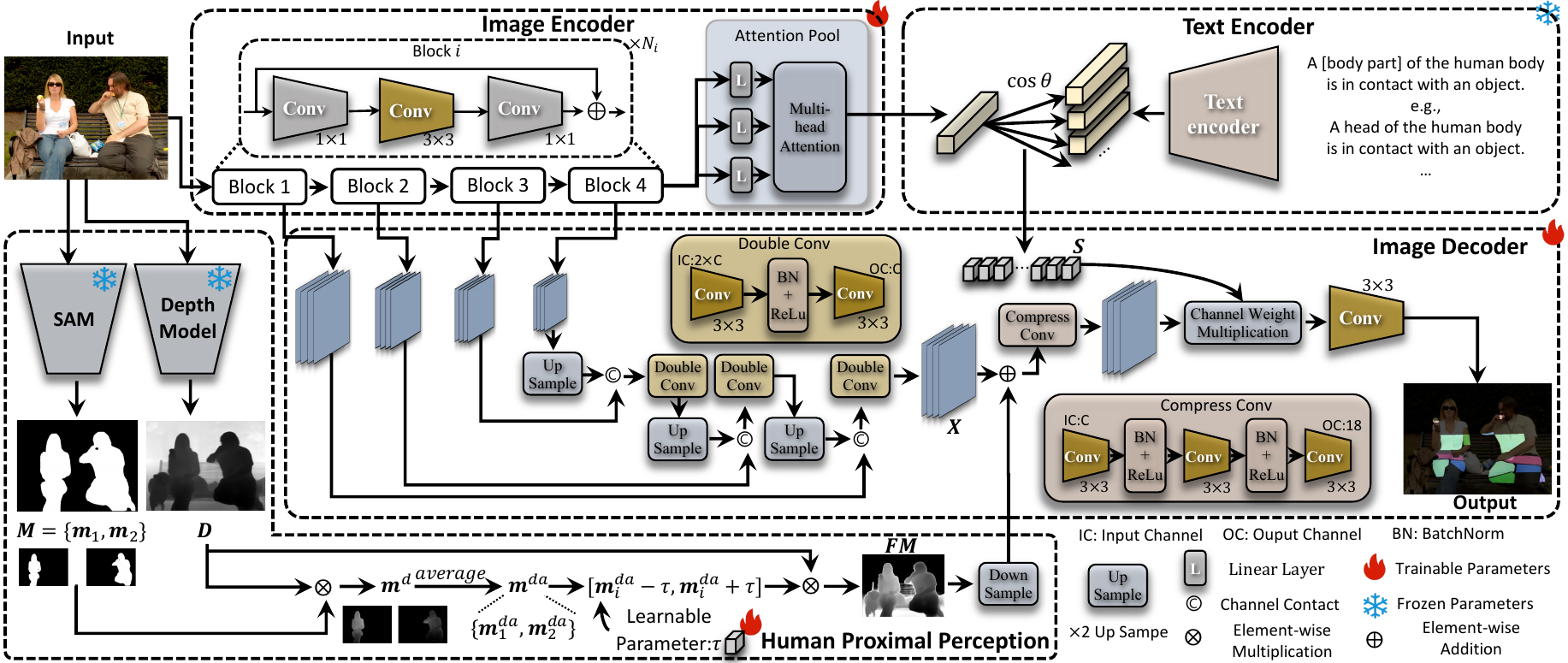}
\caption{Overall architecture. The image encoder and text encoder are used to extract features from images and text, respectively. In the text encoder, we design a prompt template to create specific textual prompts. The similarity between the image and text features is computed to enhance the attention in the image decoder. The HPP module calculates the human mask and depth features, and uses learnable parameters, $\tau$, to capture the depth range around the human body and surrounding environment. The image decoder progressively refines the segmentation features by integrating each block output from the image encoder.
}
\label{fig:network}
\end{figure*}

\textbf{HOI detection.} HOI focuses on detecting humans and objects and predicting the classification of interactions between humans and objects. Chao et al.~\cite{chao2018learning} first proposed a two-stage model using object detection and classification, in which a pre-trained detection model is initially employed to detect humans and objects, followed by a classification network to recognize interactions between them. However, the two-stage approach~\cite{chao2018learning,gkioxari2018detecting,tip-9552553} decomposes the HOI task, leading to reduce efficiency. Therefore, Zou et al.~\cite{zou2021end} slightly modified the DETR~\cite{carion2020end} network, based on a transformer architecture, to detect humans, objects, and interactions simultaneously. It is categorized as a one-stage HOI method, which is more efficient~\cite{wang2024cyclehoi,yang2024open}.
In addition, some researchers utilized the CLIP~\cite{radford2021learning} to integrate text information into networks, further enhancing accuracy~\cite{liao2022gen,wang2024ted}. 

\textbf{HOT detection.} Early HOT detection primarily focused on specific body parts such as hands and feet ~\cite{narasimhaswamy2020detecting,kobayashi2024estimation,zhuo2023towards,cui2023hand,tripathi2023deco,yang2024learning,moon2024dataset}. DRNet~\cite{narasimhaswamy2020detecting} combines Mask R-CNN~\cite{he2017mask} to learn hand localization and predict its contact state. The foot contact detection is often used to estimate body posture and joint angles~\cite{kobayashi2024estimation,zhuo2023towards}. These methods lack a holistic perspective, failing to provide a comprehensive assessment of HOT. To address this, Tripathi et al.~\cite{tripathi2023deco} proposed a human contact detection method based on 3D scenes.
However, this method suffers from low computational efficiency in 3D environments. Then two datasets are constructed based on 2D perspectives: HOT-Annotated and HOT-Generated.~\cite{hot}. 
DHOT introduced a combination of ResNet and convolutional attention mechanisms to achieve HOT predictions~\cite{hot}. 
Later, PIHOT~\cite{wang2024pihot} was designed to address the human-object occlusion problem. PIHOT employs a restoration model and a depth model to recover object features.
However, these methods do not consider category consistency within a region, leading to frequent occurrences of other categories within a specific category.
Moreover, PIHOT has a large number of parameters, runs slowly, and does not exploit multimodal information to enhance contact detection performance.


\textbf{Text-guided perception method.} Multimodal information, such as text prompts, has been proven effective in various tasks ~\cite{liao2022gen, wang2024freea, wang2024review}. GEN-VLKT~\cite{liao2022gen} leverages the CLIP model to extract text features, which are then used to initialize a fully connected layer, thereby improving accuracy. FreeA~\cite{wang2024freea} employs the CLIP model to compute similarity between images and prompt texts, enabling the generation of candidate actions for label-free tasks. OpenCat~\cite{zheng2023open} enhances model performance through the use of text prompts. In contrast, our method utilizes text prompts to guide the network's attention to 17 body parts (excluding the background), thereby improving segmentation accuracy by focusing more precisely on body parts. 


\textbf{Why choose 2D over 3D contact detection?} 
In many real-world applications, 2D tasks are sufficient for detecting HOT.
For example, when understanding human-object interactions, it is sufficient to know whether the hands, feet, or other body parts are in contact with an object, without requiring precise 3D contact information. While 3D contact detection provides more information, it requires a larger number of parameters due to the complexity of 3D models, leading to slower speed. In contrast, 2D HOT detection is more suitable for some real-world applications. Moreover, 3D HOT detection datasets are currently scarce and challenging to construct. Although some recent works have attempted to create 3D HOT detection datasets~\cite{bhatnagar2022behave,tripathi2023deco}, the available data remains limited. These datasets are typically generated from existing 2D HOT datasets using the SMPL method~\cite{tripathi2023deco}, resulting in relatively coarse annotations for contact surfaces.
Most importantly, 3D HOT detection datasets are only applicable to a limited range of objects~\cite{bhatnagar2022behave}, whereas 2D HOT datasets can be applied to a much broader variety of object categories~\cite{hot}. The number of datasets available for 2D HOT detection is gradually increasing, and their scope is also becoming broader.


\section{Method}

The proposed overall framework is illustrated in Figure \ref{fig:network}. The image is first processed by an image encoder to extract features, while simultaneously, self-constructed textual prompts are processed by a text encoder to extract text features. Next, the image and text features are matched to influence the channels of the image features, thereby introducing attention based on textual modality information prompts. A human proximal perception mechanism is designed, where the human mask is combined with depth features containing pseudo-3D information to obtain the average depth of each human. Then a learnable parameter, $\tau$, dynamically adjusts the depth range around each human, filtering out irrelevant noise regions. Additionally, to integrate fine-grained information within the feature extraction encoder, we continuously fuse features from each block of the image encoder during the decoding process. Finally, the HOT prediction map is obtained.

\subsection{Image Encoder}
\label{sec_image_encoder}

The ResNet-50 with an attention pooling layer is used as our image encoder. Given an image $\bm{I}$, each block of ResNet-50 outputs a feature map $\bm{F}_i \in \mathbb{R}^{C_i \times \frac{H}{S_i} \times \frac{W}{S_i}}$, where $ i = \{1, 2, 3, 4\} $, the channel dimensions $ C_i = \{\text{256, 512, 1024, 2048}\} $ and the downsmaple ratio of each block $ S_i = \{\text{4, 8, 16, 32}\} $. To compute similarity with text features, we flatten $ \bm{F}_4 $ into a one-dimensional vector through an attention pool to output $ \bm{F}_{IE} \in \mathbb{R}^{1 \times 1024} $. The attention pool, proposed by Radford et al.~\cite{radford2021learning}, is primarily designed to capture key information within the image and align it with textual descriptions for enhanced matching with text features.


\subsection{Text Encoder}








To extract texts features, we first construct a text prompts template $ T $, which takes the form: ``A [body part] of the human body is in contact with an object." Here, [body part] is sequentially replaced with each of the 17 human body parts (``Head", ``Chest", ``Left Upper Arm", ``Left Fore Arm", ``Left Hand", ``Right Upper Arm", ``Right Fore Arm", ``Buttocks", ``Right Hand", ``Hip", ``Back", ``Left Thigh", ``Left Calf", ``Left Foot", ``Right Thigh", ``Right Calf", ``Right Foot") mentioned earlier. After encoding, we obtain $ \bm{F}_{TE} \in \mathbb{R}^{TN\times1024}$, where $TN$ denotes the number of constructed texts. The text encoder is initialized using the model proposed by Radford et al.~\cite{radford2021learning}, freezing its parameters during training. The similarity $ \bm{S} \in \mathbb{R}^{1 \times TN} $ between $ \bm{F}_{IE} \in \mathbb{R}^{1 \times 1024} $ and $ \bm{F}_{TE} \in \mathbb{R}^{TN\times1024} $ is then computed, via:

\begin{equation}
    \bm{S} = \frac{\bm{F}_{IE} \cdot {\bm{F}_{TE}}^\text{T}}{\|\bm{F}_{IE}\| \cdot \|\bm{F}_{TE}\|}.
\end{equation}

Since we train the image encoder while keeping the text encoder frozen, the two would quickly become misaligned without an image-text similarity loss constraint. Therefore, after $\bm{S}$, we compute the loss with the ground truth to further update the image encoder parameters, ensuring alignment with the text encoder and preventing conflicts.

\subsection{Human Proximal Perception}

This module aims to preserve humans and the environment around them by eliminating irrelevant backgrounds. To accomplish this, the initial step is calculating the average depth for each person. Then, the depth range around each person is determined based on the learnable parameter $ \tau $ to dynamically select the appropriate target region.
Specifically, the improved SAM model is used to generate humans masks $ \bm{M} $ based on the text prompt ``person"~\cite{github-repo}, and apply the ZoeDepth model~\cite{bhat2023zoedepth} to extract the depth features $ \bm{D}\in \mathbb{R}^{H \times W} $ of the input image. Here, $ \bm{M} = \{ \bm{m}_1, \bm{m}_2, \dots, \bm{m}_N \} $, where $ \bm{m}_i \in \mathbb{R}^{H \times W} $ represents the $ i $-th human mask. In $ \bm{m}_i $, 0 indicates the background, and 1 represents the human body. $ N $ denotes the total number of persons. Next, we obtain the average depth of each human based on $ \bm{M} $ and $ \bm{D} $ as follows:
\begin{equation}
\label{eq:norm}
    \bm{D}_{Norm} = \frac{\bm{D} - \text{Min}(\bm{D})}{\text{Max}(\bm{D}) - \text{Min}(\bm{D})},
\end{equation}
\begin{equation}
    \bm{m}^d_i = \bm{m}_i \otimes \bm{D}_{Norm}, i=1,2,\dots,N,
\end{equation}
\begin{equation}
    \bm{m}^{da}_i = \frac{\sum_{h=1}^{H} \sum_{w=1}^{W} \bm{m}^{d}_{i}[h, w]}{\sum_{h=1}^{H} \sum_{w=1}^{W} \bm{m}_{i}[h, w]},
\end{equation}
where $\otimes$ represents element-wise multiplication. Eq.~\ref{eq:norm} is used to normalize $ \bm{D} $ within the range $[0,1]$, \text{Min} and \text{Max} denote the minimum and maximum values, respectively. $ \bm{m}^d_i $ represents the depth matrix of the $ i $-th human, and $ \bm{m}^{da}_i$ denotes the average depth value for $ i $-th human. Subsequently, based on the learned parameter $ \tau $, we obtain the depth range $ \bm{d}^r $ via:
\begin{equation}
    \bm{d}^r_i = \{ [ \bm{m}^{da}_i - \tau, \bm{m}^{da}_i + \tau ] | i=1,2,\dots,N \}.
\end{equation}


Next, $ \bm{d}^r $ is used to generate a filter mask matrix $ \bm{FM} \in \mathbb{R}^{H \times W} $, which is composed only of 0 and 1, to retain the valid regions of the human and surrounding objects. That is:
\begin{align}
\bm{FM} & = \begin{bmatrix}  
  a_{11}& a_{12}& \cdots  & a_{1W} \\  
  a_{21}& a_{22}& \cdots  & a_{2W} \\  
  \vdots & \vdots & \ddots & \vdots \\  
  a_{H1}& a_{H2}& \cdots  & a_{HW}  
\end{bmatrix}_{H\times W} & = \left [ a_{pq}\right ],
\end{align}
\begin{align}
     a_{pq} &=\begin{cases}1,\text{ if } \bm{m}^{da}_i - \tau<\bm{D}^{pq}_{Norm}<\bm{m}^{da}_i + \tau,
 \\0,\text{ otherwise },
\end{cases}
\end{align}
where $p$ and $q$ represent the element indices at the $p$-th row and $q$-th column of the matrix. $ \bm{FM} $ will be downsampled and added with the feature map $\bm{X}$ from the image decoder, thereby highlighting the human body and its surrounding regions. However, since $ \bm{FM} $ is generated through a comparison operation that directly assigns elements to 0 or 1, it is not a continuous function, and thus, it cannot be backpropagated, meaning the parameter $ \tau $ cannot be updated. To address this, we propose an alternative solution as follows:
\begin{equation}
\bm{FM} = [0]_{H\times W},
\end{equation}
\vspace{-0.7cm}
\begin{equation}
\label{eq:Theta}
    \bm{\Theta_i } = (\bm{D}_{Norm} - (\bm{m}^{da}_i - \tau)) \otimes ((\bm{m}^{da}_i + \tau)) - \bm{D}_{Norm})),
\end{equation}
\vspace{-0.5cm}
\begin{equation}
\label{eq:FM}
    \bm{FM} = \bm{FM} + \sum_{i=1}^{N} \text{ReLU}(\bm{\Theta_i } ),
\end{equation}
\begin{equation}
\label{eq:FMxX}
        \bm{O} = \text{DS}(\bm{FM}) \oplus \bm{X}.
\end{equation}


$[0]_{H \times W}$ denotes the initialization of a zero matrix of size $H \times W$. Eq.~\ref{eq:Theta} converts the depth values within the specified range to positive values, while those outside the range are set to negative values. The ReLU function in Eq.~\ref{eq:FM} keeps the original positive values and turns the negative values into zeros. Since the ReLU function is differentiable, it allows the parameter \( \tau \) to be updated during backpropagation. \text{DS} denotes the downsampling operation, which reduces the size of $\bm{FM}$ from $H\times W$ to $\frac{H}{4}\times \frac{W}{4}$. $\oplus$ represents element-wise addition. $\bm{O}$ denotes the fused feature map.

\begin{figure*}[ht]
    \centering
    \includegraphics[width=1.0\linewidth]{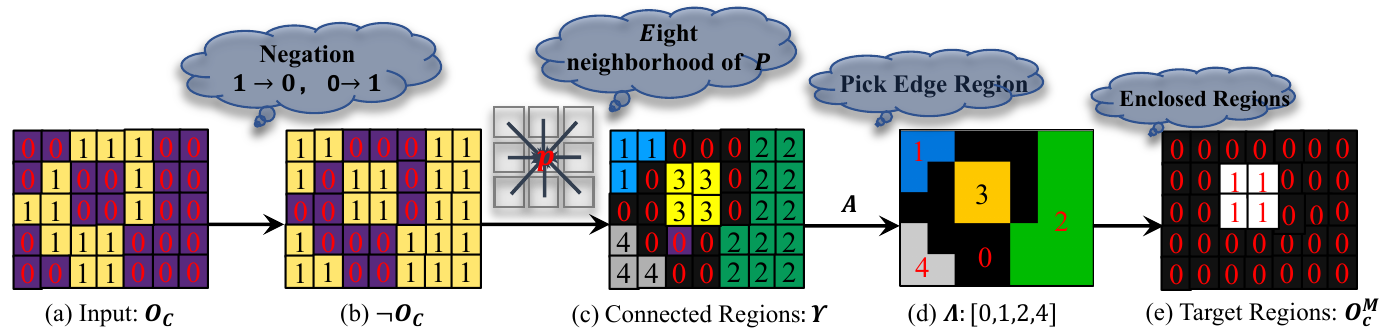}
    \caption{The calculation schematic of regions of other categories enclosed within category \( c \). First, the input matrix values equal to \( c \) are set to 1, while all other values are set to 0, forming the input \( \bm{O}_c \) as shown in subfigure (a). Then, \( \bm{O}_c \) is inverted to obtain \( \neg \bm{O}_c \). The connected regions of \( \neg \bm{O}_c \) are computed, resulting in \( \bm{\Upsilon} \). Next, the values from the first and last rows and columns of \( \bm{\Upsilon} \) are extracted, forming the set \( \bm{\Lambda  } = \{0,1,2,4\} \). Since the value 3 is missing in \( \bm{\Lambda} \), we set all elements in  \( \bm{\Upsilon} \) that are equal to 3 to 1 and the rest to 0, yielding $\bm{O}_c^M$. The region in \( \bm{O}_{c}^M \) where elements are 1 represents the area of other categories enclosed within category \( c \).
}
    \label{fig:gjl}
\end{figure*}

\subsection{Image Decoder}


The decoder network performs feature decoding. Specifically, it first upsamples the outputs from the four blocks of the encoder in sequence to fuse fine-grained features. Then, after integrating features from the HPP layer, it connects them through the attention output of the text encoder to predict the final segmentation map.

After extracting the features in Sec. \ref{sec_image_encoder}, it is necessary to continually upsample them to create the segmentation map. In the image encoder, the input image is continuously downsampled, resulting in a gradual loss of image details. Thus, when upsampling, it is necessary to combine the results $\bm{F}$ from the less deep blocks in the encoder network in order to regain missing details like image textures $\bm{X} \in \mathbb{R}^{256\times \frac{H}{4} \times \frac{W}{4}}$, via:
\begin{equation} \nonumber
    \bm{x}_4 = \bm{F}_4,
\end{equation}
\vspace{-0.7cm}
\begin{equation} \nonumber
    \bm{x}_{i-1} = \text{Double Conv}(Up(\bm{x}_i) \textcircled{c} \bm{F}_{i-1}), i=4,3,2,
\end{equation}
\vspace{-0.5cm}
\begin{equation}
    \bm{X} = \bm{x}_1,
\end{equation}
where \textcircled{c} denotes channel concatenation. Subsequently, the features $\bm{X}$ are processed through Eq.\ref{eq:FMxX} to obtain $\bm{O}$, focusing solely on the human and the surrounding area of human within a certain range. To focus on the text input, \text{Compress Conv} initially reduces the channels of $\bm{O}$ to 18 before combining them using the equation below:
\begin{equation} \nonumber
    \bm{O} = \text{Compress Conv}(\bm{O}),
\end{equation}
\begin{equation}
    \bm{O}[i] = \begin{cases}{\bm{O}}[i] \times 1,\text{ if } i=18
 \\\bm{O}[i] \times \bm{S}[i],\text{ if } i=1,2,\dots,17,
\end{cases}
\end{equation}
where $\bm{O}[i]$ represents the background feature map based on channel dimension when $i$ is 18. We keep the background feature map unchanged and adjust the remaining channels based on $\bm{S}$. Finally, $\bm{O}$ is passed through the convolution layers to obtain the final segmentation map $\bm{O} \in \mathbb{R}^{18 \times \frac{H}{4} \times \frac{W}{4}}$. It is not difficult to observe that the output of the network, $\bm{O}$, does not have a dimension of $18 \times H \times W$, but instead undergoes a 4$\times$ downsampling. This is because we also apply a 4$\times$ downsampling to the ground truth ($\bm{GT} \in \mathbb{R}^{\frac{H}{4} \times \frac{W}{4}}$), which is consistent with DHOT~\cite{hot} and PIHOT~\cite{wang2024pihot}.


\subsection{Loss}


HOT categorizes the areas of contact between humans and objects into 18 classes, which include the background. Current techniques that utilize cross-entropy loss frequently produce segmentation outcomes that contain inaccurate categories in specific regions of the resultant map.
To tackle this problem, we create a Regional Joint Loss, known as RJLoss. RJLoss is made up of two parts: Local Joint Loss and Global Joint Loss.

\textbf{Local Joint Loss}. Initially, we utilize the ground truth ($\bm{GT}$) to isolate regions specific to each class. If there are additional classes present in a region, the loss for that region is adjusted accordingly. Specifically, the Local Joint Loss for a given class $c$ is defined as:
\begin{equation} \nonumber
    \bm{O}_c(\bm{GT}_{c})=\{o_{pq}(gt_{pq})|p,q=1,2,\dots,\frac{H(W)}{4} \},
\end{equation}
\begin{equation}
    o_{pq}(gt_{pq})=\begin{cases}
  1,& \text{ if } o_{pq}(gt_{pq})=c \\
  0,& \text{ otherwise }
\end{cases},
\end{equation}
\begin{equation}
\label{eq:l_c}
    \mathcal{L}^L_c=\frac{\sum (|\bm{O}_c-\bm{GT}_c| \otimes \bm{GT}_c)}{\sum \bm{GT}_c },
\end{equation}
where $|\cdot|$ denotes the absolute value. In Eq.\ref{eq:l_c}, $|\bm{O}_c - \bm{GT}_c|$ sets the parts where $\bm{O}_c$ equals $\bm{GT}_c$ to 0 and the parts where they differ to 1. The $|\bm{O}_c - \bm{GT}_c| \times \bm{GT}_c$ identifies cases where other classes appear within class $c$. The symbol $\sum$ denotes the summation over all elements of the matrix. Subsequently, the Local Joint Loss for all classes is defined as:
\begin{equation}
    \mathcal{L}^L=\sum_{c=1}^{18} \mathcal{L}^L_c
\end{equation}

\textbf{Global Joint Loss}. The Local Joint Loss aims to eliminate classes from the target class region that do not belong there, as determined by the $\bm{GT}$. However, it fails to consider the entire prediction map, including non-contact areas of the human body. The Global Joint Loss calculates the joint loss over the entire prediction map. First, we need to identify the connected regions of class $c$ and compute the regions of other classes enclosed within them (the simplified schematic diagram is shown in Figure \ref{fig:gjl}(a), (b), (c) and (d)) using
\begin{equation}\nonumber
    \bm{A}=[a_{ij}]=\begin{pmatrix}  
  1 & \cdots & 1 \\  
  \vdots & 0& \vdots \\  
  1 & \cdots & 1  
\end{pmatrix}_{\frac{H}{4} \times \frac{W}{4}  },
\end{equation}
\begin{equation}\nonumber
    \bm{\Upsilon} = \text{ConnectedArea}(\neg  \bm{O}_c),
\end{equation}
\begin{equation}
    \bm{\Lambda  }=\{\bm{\Upsilon}_{ij}|\forall i,j\in[1,\frac{H}{4} ],[1,\frac{W}{4} ] \text{ and } a_{ij}=1\},
\end{equation}
where $\bm{A}$ represents a matrix with boundaries set to 1 and the interior set to 0, and $\neg$ denotes the negation operation on the matrix, flipping 0 to 1 and 1 to 0 (as illustrated in Figure \ref{fig:gjl}(a) and (b)). \text{ConnectedArea} is a function for finding connected regions, implemented using the built-in functions from the scipy library. It detects all connected regions in the matrix and assigns a unique label to each region (as illustrated in Figure \ref{fig:gjl}(c)). Specifically, it iterates over all elements in the input matrix, searching in 8 directions (up, down, left, right, top-left, bottom-left, top-right, bottom-right) for values equal to the target element, and marks them as a connected region. $\bm{\Lambda}$ denotes several unequal region labels extracted from the boundary of the $\bm{\Upsilon}$ through $\bm{A}$ (as shown in Figure \ref{fig:gjl}(d)). Then the connected regions are identified not in $\bm{\Lambda}$ (as shown in Figure \ref{fig:gjl}(e)) using the following formula, which gives us the target regions enclosed by contact class $c$.
\begin{equation}\nonumber
    \bm{O}^M_c = [m_{ij}]
\end{equation}
\begin{equation}
    m_{ij}=\begin{cases}
  1,& \text{ if } \bm{\Upsilon}_{ij} \notin \Lambda \\
  0,& \text{otherwise}
\end{cases}.
\end{equation}

Then, the internal error class loss is calculated based on $\bm{O}^M_c$:
\begin{equation}
    \mathcal{L}^G_c = \sum (\neg \bm{O}_c\otimes \bm{O}^M_c).
\end{equation}

The Global Joint Loss for all classes is defined as:
\begin{equation}
    \mathcal{L}^G = \sum_{c=1}^{18} \mathcal{L}^G_c.
\end{equation}


The overall network loss is optimized jointly using cross-entropy and RJLoss Additionally, a basic image-text matching loss is incorporated to co-optimize the image encoder. The total loss is defined as:
\begin{equation}
\label{eq:total_loss}
    \mathcal{L} = \text{CE}(\bm{O}, \bm{GT}) + \alpha \mathcal{L}^L + \beta \mathcal{L}^G + \gamma \text{BE}(\bm{S},\bm{\mathbb{C}}),
\end{equation}
where \text{CE} and \text{BE} denote the cross-entropy loss and the binary cross-entropy loss, respectively. Both the image and text encoders are initialized with
CLIP. We freeze the text encoder and train the image encoder using \text{BE}.
$\bm{S} \in \mathbb{R}^{1 \times 18}$ represents the image-text matching similarity, and $\bm{\mathbb{C}} \in \mathbb{R}^{1 \times 18}$ indicates the contact classes present in the input image. If class $c$ appears in the image, then $\bm{\mathbb{C}}_c = 1$; otherwise, $\bm{\mathbb{C}}_c = 0$.



\section{Experiment}

\subsection{Datasets}
To assess the performance of the proposed approach, two benchmark datasets, HOT-Annotated and HOT-Generated, are employed. The HOT-Annotated comprises 15,082 images with 67,088 contact areas from V-COCO~\cite{gupta2015visual}, HAKE~\cite{li2020pastanet}, and Watch-n-Patch~\cite{wu2015watch}. The HOT-Generated dataset is constructed by using the PROX~\cite{hassan2019resolving} and SMPL-X~\cite{pavlakos2019expressive} frameworks, featuring 20,205 images and a total of 95,179 contact areas.

\subsection{Setup}
For a fair comparison, ResNet-50 is used as the backbone of the image encoder module. $\alpha$, $\beta$, and $\gamma$ are set to 0.3, 0.1, and 1.0, respectively. The network is optimized using the AdamW optimizer. The batch size is set to 4 per GPU.
The experimental environment is Ubuntu 20.04, equipped with 8 NVIDIA A6000 GPUs. PyTorch version is 1.11.0, torchvision version is 0.12.0, and Python is 3.8.19.


\subsection{Metrics}

\begin{figure*}[ht]
\centering 
\includegraphics[width=\linewidth]{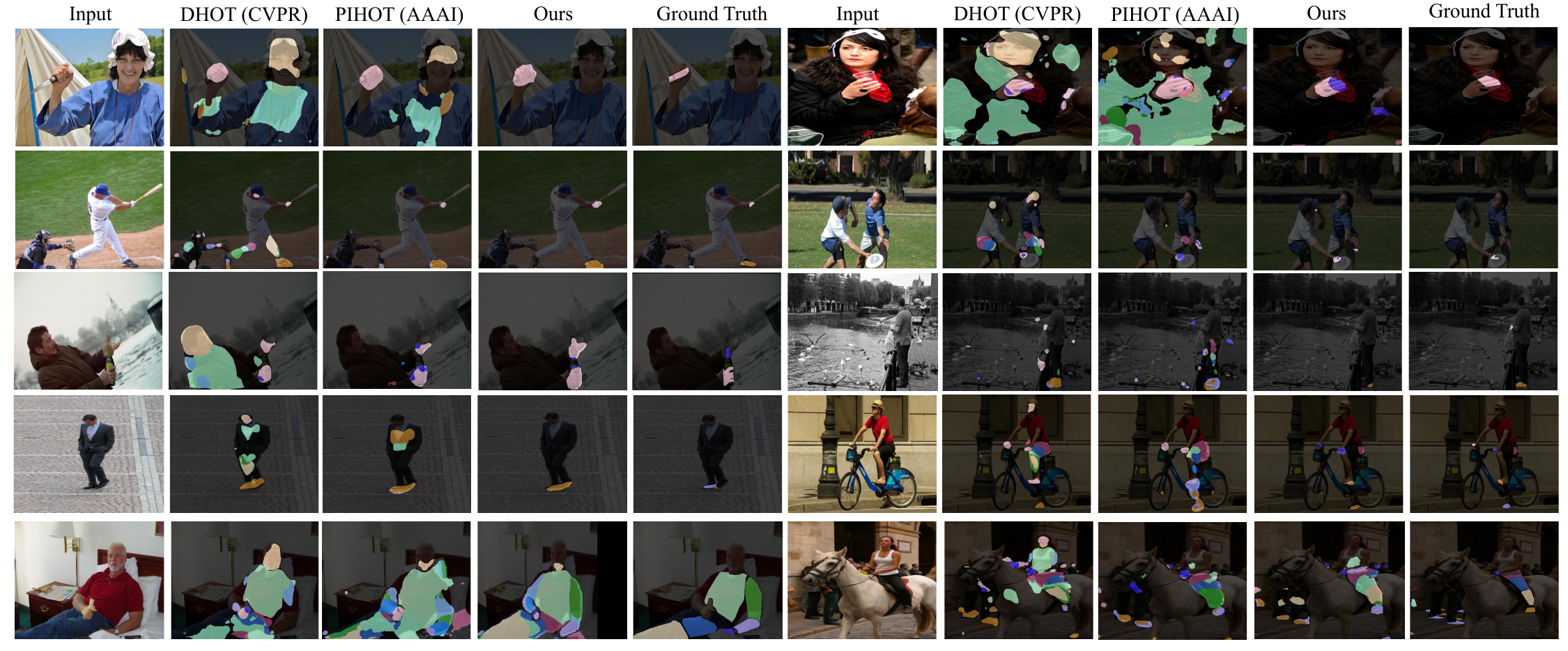}
\caption{Visualization of DHOT~\cite{hot}, PIHOT~\cite{wang2024pihot}, and our proposed method.
}
\label{fig:vis_reuslts}
\end{figure*}


\begin{table*}[]
\centering
\small
\begin{threeparttable}
\begin{tabular}{l|ccccc|ccccc}

\specialrule{1.5pt}{0pt}{0pt} 
\multirow{2}{*}{Model} & \multicolumn{5}{c|}{HOT-Annotated}         & \multicolumn{5}{c}{HOT-Generated}          \\
                       & SC-Acc. & C-Acc. & mIoU  & wIoU  & AD-Acc. & SC-Acc. & C-Acc. & mIoU  & wIoU  & AD-Acc. \\ \hline
ResNet+UperNet~\cite{xiao2018unified}         & 35.1    & 62.6   & 19.5 & 22.7 & -       & 21.1    & 42.7   & 8.0 & 11.6 & -       \\
ResNet+PPM~\cite{zhao2017pyramid}             & 34.6    & 61.1   & 20.1 & 23.3 & -       & 21.2    & 41.1   & 7.5 & 11.9 & -       \\
DHOT(ResNet-50)$_\text{wo/att}$~\cite{hot}                    & 24.1    & 42.8   & 14.8 & 18.7 & -       & 12.0    & 24.6   & 5.1 & 9.9 & -       \\
DHOT(ResNet-50)$_\text{pure\_att}$~\cite{hot}                    & 33.8    & 58.4   & 18.9 & 23.7 & -       & 20.3    & 40.1   & 7.7 & 11.3 & -       \\
DHOT(ResNet-50)$_\text{Full}$~\cite{hot}                    & 40.7    & 70.7   & 21.5 & 26.0 & -       & 30.4    & 54.3   & 13.9 & 16.7 & -       \\
DHOT(ResNet-50)$_\text{Full}$-OF~\cite{hot}                    & 40.1    & 69.2   & 22.1 & 25.0 & 31.0   & 24.3      & 59.2     & 12.0    & 13.0    & 24.9      \\
PIHOT(ResNet-50)~\cite{wang2024pihot} & 45.3    & \textbf{80.7}   & 23.6 & 28.6 & 31.3   & 34.9      & \textbf{76.3}     & 16.9    & 21.2    & 25.4 \\
Ours(ResNet-50)                   & \textbf{46.0}   & 74.9  & \textbf{25.6} & \textbf{30.2} & \textbf{42.3}   & \textbf{35.2}      & 70.8     & \textbf{18.0}    & \textbf{23.1}    & \textbf{30.6}      \\ 
\specialrule{1.5pt}{0pt}{0pt} 
\end{tabular}
\begin{tablenotes}
\footnotesize
\item[*] -OF indicates the optimal result provided by themselves in Github.
\end{tablenotes}
\end{threeparttable}
\caption{Performance comparisons on HOT-Annotated and HOT-Generated datasets.}
\label{tab:results}
\end{table*}

HOT was proposed in 2023 by Chen et al.~\cite{hot}, and it includes four evaluation metrics: SC-Acc., C-Acc., mIoU, and wIoU. SC-Acc. represents the proportion of correctly classified pixels. C-Acc. denotes the accuracy of classifying pixels on the human body, which is a binary classification. mIoU measures the Intersection over Union between the predicted and ground truth regions, while wIoU is the weighted IoU across all classes. To adjust mIoU and wIoU to the same numerical range as SC-Acc. and C-Acc., i.e., 0-100, we multiply mIoU and wIoU by 100.

We found that the C-Acc. metric has some issues. If the prediction map classifies the entire image as a contact class, C-Acc. would be 100\% because the predicted region encompasses the entire human body. This outcome is clearly undesirable, as most of the prediction pixels are incorrect in this case. To address this issue, we propose a new evaluation metric to replace C-Acc., called Adaptive Accuracy (AD-Acc.). We first extract the channel indices with the maximum values from the prediction map, resulting in a new $\bm{O} \in \mathbb{R}^{\frac{H}{4} \times \frac{W}{4}}$, where each element represents the predicted class. The calculation of AD-Acc. is then defined as follows:
\begin{align}\nonumber
    \bm{GT}^B(\text{ or }\bm{O}^B) &= [gt_{ij}](\text{ or }[o_{ij}]) \\
    &= \begin{cases}
  1,&\text{if } gt_{ij} (\text{ or } o_{ij}) >0 \\
  0,&\text{otherwise}
\end{cases},
\end{align}
\begin{equation}
    \bm{\zeta } = \bm{M} - \bm{GT}^B \otimes \bm{M},
\end{equation}
\begin{equation}
    \text{AD-Acc.} = \frac{\sum (\bm{GT}^B \otimes \bm{O}^B)}{\sum \bm{GT}^B + \delta } - \frac{\sum(\bm{\zeta } \otimes \bm{O}^B))}{\sum \bm{\zeta } + \delta},
\end{equation}
where $\bm{GT}^B$ and $\bm{O}^B$ denote the binarized of $\bm{GT}$ and $\bm{O}$, respectively. $\bm{M}$ represents the human mask. $\bm{\zeta}$ refers to the negative samples selected based on the human mask, specifically the parts of the human body excluding the ground truth. $\delta$ is set to 1e-6. $\text{AD-Acc.}$ takes into account both positive and negative samples, balancing cases where human classification errors occur, making it a more robust evaluation metric.

\subsection{Performance}
The experimental results (Table \ref{tab:results}) demonstrate that our proposed model significantly outperforms all other models on both HOT-Annotated and HOT-Generated datasets across four evaluation metrics. For example, on the HOT-Annotated dataset, our model achieves a SC-Acc. of 46.0, which is notably higher than PIHOT at 45.3 and DHOT(ResNet-50)$_\text{Full}$-OF at 40.1. Similarly, for mIoU, our model scores 25.6 compared to PIHOT at 23.6 and DHOT(ResNet-50)$_\text{Full}$-OF at 22.1. Although our method is 5.8 points lower than PIHOT in terms of C-Acc., we previously mentioned that C-Acc. is not an accurate metric for evaluating the HOT task. This is because if every pixel in the entire image is predicted as a single contact category, C-Acc. would be 100, which is clearly not the desired outcome. To address this limitation, we propose a new evaluation metric, AD-Acc. In terms of AD-Acc., our method outperforms PIHOT by 11.0. On the HOT-Generated dataset, our model achieves a AD-Acc. of 30.6, surpassing PIHOT at 25.4 and DHOT(ResNet-50)$_\text{Full}$-OF at 24.9. These substantial improvements emphasize the effectiveness of our approach and the critical role of fully integrated various mechanisms in achieving superior performance for HOT prediction tasks. Compared to PIHOI, our method is more computationally efficient, has fewer parameters, and lays the foundation for incorporating more modalities.

\subsection{Ablation Study}

\textbf{Various components}. In Table \ref{tab:vc}, all experiments were conducted using only cross-entropy loss to train the network model, with $\alpha$, $\beta$, and $\gamma$ all set to 0. The baseline refers to using only the image encoder and image decoder, where the decoder directly performs convolution and prediction on the features from the last layer of the encoder. +Fine denotes refining features by combining the output of each block from the encoder during upsampling in the image decoder.
TE and DE denote the test encoder and depth model, respectively. ``+DM+SAM'' denotes the HPP module. The results clearly demonstrate the individual contribution of each component as well as their combined impact.

\begin{table}[h]
\centering
\small
\setlength{\tabcolsep}{1.1mm}{
\begin{tabular}{l|ccccc}
\specialrule{1.5pt}{0pt}{0pt} 

         & SC-Acc. & C-Acc. & mIoU  & wIoU  & AD-Acc. \\ \hline
Baseline (BA) & 37.2    & 65.9   & 19.0 & 22.9 & 30.3       \\
BA +Fine (BF)    & 38.9    & 68.4   & 20.1 & 24.5 & 34.1        \\
BF +TE      & 40.3    & 69.6   & 21.0 & 25.9 & 37.1       \\
BF+DM & 40.1    & 68.2   & 20.4 & 24.9 & 36.5 \\
BF+SAM & 39.8    & 67.9   & 20.1 & 24.4 & 35.3 \\
BF+TE+DM & 41.6    & 70.1   & 22.4 & 26.7 & 37.5 \\
BF+TE+SAM & 41.3    & 69.2   & 21.9 & 26.0 & 37.4 \\
BF+TE+DM+SAM & \textbf{43.2}   & \textbf{71.8}  & \textbf{23.1} & \textbf{27.7} & \textbf{38.9} \\
\specialrule{1.5pt}{0pt}{0pt} 

\end{tabular}}
\caption{Ablation experiments of adding various components on HOT-Annotated.}
\label{tab:vc}
\end{table}

\textbf{Different loss}. The impact of adding different loss functions on the experimental results is listed in Table \ref{tab:dlf}. CE denotes using only cross-entropy loss. +BE indicates the addition of binary cross-entropy loss on CE. +RJLoss represents further incorporation of the Regional Joint Loss proposed in this paper. By introducing the text modality and optimizing with BE, the performance improves. After adding RJLoss, the optimal results are achieved, with a significant performance boost.


\begin{table}[h]
\centering
\small
\begin{tabular}{l|ccccc}
\specialrule{1.5pt}{0pt}{0pt} 

         & SC-Acc. & C-Acc. & mIoU  & wIoU  & AD-Acc. \\ \hline
CE & 43.2    & 71.8   & 23.1 & 27.7 & 38.9       \\
+BE    & 44.5    & 72.3   & 23.8 & 28.3 & 40.1        \\
+RJLoss    & \textbf{46.0}   & \textbf{74.9}  & \textbf{25.6} & \textbf{30.2} & \textbf{42.3}
\\ 
\specialrule{1.5pt}{0pt}{0pt} 

\end{tabular}
\caption{Impact of different loss functions on performance.}
\label{tab:dlf}
\end{table}

\textbf{The range of $\bm{D}$}. The depth map is used to adaptively select the depth around the human body based on the human mask and the learnable parameter $\tau$. The $\tau$ is influenced by the range of the depth map. The performance of normalized and non-normalized depth maps are compared, as shown in Table \ref{tab:dm}, where $R$ denotes real numbers. When the range of $\bm{D}$ is unconstrained, different input images have varying depths, making it difficult to optimize $\tau$ within a specific interval, resulting in lower performance. After normalizing $\bm{D}$, the results reached the optimal performance.


\begin{table}[h]
\centering
\small
\begin{tabular}{l|ccccc}
\specialrule{1.5pt}{0pt}{0pt} 

         & SC-Acc. & C-Acc. & mIoU  & wIoU  & AD-Acc. \\ \hline
$\bm{D} \in R$ & 44.1    & 72.4   & 24.3 & 28.1 & 40.2       \\
$\bm{D} \in [0,1]$    & \textbf{46.0}   & \textbf{74.9}  & \textbf{25.6} & \textbf{30.2} & \textbf{42.3}
\\ 
\specialrule{1.5pt}{0pt}{0pt} 

\end{tabular}
\caption{Performance comparison of depth map $\bm{D}$ Range.}
\label{tab:dm}
\end{table}

\textbf{Different loss weights}. The impact of different loss weights on performance is listed in Table \ref{tab:dlw}. When $\alpha$, $\beta$, and $\gamma$ are all set to 1.0, the performance is significantly reduced. We observed that the early values of $\mathcal{L}^L$ and $\mathcal{L}^G$ are quite large, causing the model to overlook the original segmentation task and instead focus excessively on the presence of other classes within regions. Consequently, as $\alpha$ and $\beta$ decrease, the model's performance gradually improves, achieving optimal results at $\alpha = 0.3$ and $\beta = 0.5$.


\begin{table}[h]
\small
\setlength{\tabcolsep}{1.8mm}{
\begin{tabular}{ccc|ccccc}
\specialrule{1.5pt}{0pt}{0pt}
$\alpha$   & $\beta$  & $\gamma$   & SC-Acc. & C-Acc. & IoU & mIoU & AD-Acc. \\ \hline
1.0 & 1.0 & 1.0 &  42.3       &   71.9     &  23.1   &   27.7   & 40.1   \\
0.3 & 0.1 & 1.0 & \textbf{46.0} & \textbf{74.9} & \textbf{25.6} & \textbf{30.2} & \textbf{42.3}   \\
0.3 & 0.1 & 0.5 &   45.5      &  73.8      &   24.9  &   29.2   & 41.7   \\
\specialrule{1.5pt}{0pt}{0pt}
\end{tabular}}
\caption{Effect of different loss weights on performance.}
\label{tab:dlw}
\end{table}

\subsection{Visualization}
In Figure \ref{fig:vis_reuslts}, 
it is evident that our proposed method aligns closely with the ground truth. The discrepancy between DHOT’s results and the actual scene is particularly noticeable. 
Similarly, PIHOT produces poor results in certain cases, such as the second sample in the first row and the first sample in the last row. 
In contrast, our framework refines predictions through progressive upsampling and utilizes depth information to locate areas around the human body, suppressing irrelevant regions outside the body. Furthermore, by optimizing with RJLoss, our approach better focuses on contact areas and ensures category consistency across the entire region. 

\section{Conlusion}
This paper introduces a prompt guidance and human proximal perception method for HOT prediction (P3HOT). To enhance the performance of models that utilize only one type of image, prompt guidance are being used for the first time to help the network focus more on human body contact areas. Additionally, a human proximal perception mechanism is employed to dynamically perceive key depth information around the human body based on learnable parameters, excluding areas where interactions are improbable. In the decoder, features are gradually combined during upsampling to improve segmentation boundaries and restore initial texture details. Importantly, a new loss function named Regional Joint Loss is presented to maintain the consistency of categories within regions and reduce abnormal categories. Extensive experiments have shown that our model outperforms existing models on newly designed evaluation metrics and achieves state-of-the-art performance in SC-Acc., mIoU, and wIoU metrics on two benchmark datasets.

\section{Acknowledgments}
This work was supported in part by the National Natural Science Foundation of China under Grant 62202174, inpart by the GJYC program of Guangzhou under Grant 2024D01J0081, in part by the ZJ program of Guangdong under Grant 2023QN10X455, in part by The Taihu Lake Innovation Fund for the School of Future Technology of South China University of Technology under Grant 2024B105611004, and in part by the Guangdong Provincial Key Laboratory of Human Digital Twin (2022B1212010004).

{
    \small
    \bibliographystyle{ieeenat_fullname}
    \bibliography{main}
}

\end{document}